\documentclass[runningheads]{llncs}
\usepackage[T1]{fontenc}
\usepackage{graphicx}
\usepackage[pagebackref,breaklinks,colorlinks]{hyperref}
\usepackage[normalem]{ulem} 
\usepackage{caption}
\usepackage{subcaption}
\usepackage{algorithm2e}
\RequirePackage{amsmath}
\RequirePackage{amssymb}
\RequirePackage{booktabs}
\usepackage{multicol}
\usepackage{array}
\usepackage{multirow}
\AtEndPreamble{
    \usepackage[capitalize]{cleveref}
    \crefname{section}{Sec.}{Secs.}
    \Crefname{section}{Section}{Sections}
    \Crefname{table}{Table}{Tables}
    \crefname{table}{Tab.}{Tabs.}
}

\RequirePackage{enumitem}
\setlist[itemize]{noitemsep,leftmargin=*,topsep=0em}
\setlist[enumerate]{noitemsep,leftmargin=*,topsep=0em}

\makeatletter
\DeclareRobustCommand\onedot{\futurelet\@let@token\@onedot}
\def\@onedot{\ifx\@let@token.\else.\null\fi\xspace}

\makeatother

\begin{document}
\title{An Experimental Study on Decomposition-Based Deep Ensemble Learning for Traffic Flow Forecasting}
%

\author{
Qiyuan Zhu \inst{1} \and
A. K. Qin \inst{1}  \and
Hussein Dia \inst{1} \and
Adriana-Simona Mihaita \inst{2} \and
Hanna Grzybowska \inst{3} 
}

\authorrunning{Q. Y. Zhu, A. K. Qin et al.}
%
\institute{Swinburne University of Technology, Melbourne, Australia \ \email{\{qiyuanzhu,kqin,hdia\}@swin.edu.au} 
\and
University of Technology Sydney, Sydney, Australia \ \email{\{adriana-simona.mihaita\}@uts.edu.au} 
\and
Data61 CSIRO, Sydney, Australia \ \email{hanna.grzybowska@data61.csiro.au} }
\maketitle              
\begin{abstract}

Traffic flow forecasting is a crucial task in intelligent transport systems. Deep learning offers an effective solution, capturing complex patterns in time-series traffic flow data to enable the accurate prediction. However, deep learning models are prone to overfitting the intricate details of flow data, leading to poor generalisation. Recent studies suggest that decomposition-based deep ensemble learning methods may address this issue by breaking down a time series into multiple simpler signals, upon which deep learning models are built and ensembled to generate the final prediction. However, few studies have compared the performance of decomposition-based ensemble methods with non-decomposition-based ones which directly utilise raw time-series data. This work compares several decomposition-based and non-decomposition-based deep ensemble learning methods. Experimental results on three traffic datasets demonstrate the superiority of decomposition-based ensemble methods, while also revealing their sensitivity to aggregation strategies and forecasting horizons.

 
\end{abstract}

\section{Introduction}
Traffic flow forecasting is one of its most important tasks for intelligent transport systems (ITS) in daily traffic management and operations \cite{ref1}. Several operations, such as incident management, require reliable flow forecasting for a short horizon in future to support decision-making. However, accurate forecasting remains challenging due to the complex patterns in time-series traffic data. Factors like road congestion, vehicle breakdowns and traffic signal timing \cite{ref2,ref3} contribute to irregular and unpredictable patterns in traffic data, making accurate forecasting difficult to achieve. 

Traditional statistical and shallow machine learning techniques often struggle to capture the complex patterns within this data, rendering them less effective in this context. In contrast, deep learning techniques are better suited to adapt to the intricate nature of time-series traffic data \cite{ref4}. However, these methods can be biased, overfitting the intricate details of flow data, leading to poor generalisation.

Ensemble learning \cite{ref6} is a potential solution to mitigate the limitations of deep learning by combining the outputs of multiple models \cite{ref9}. Recent advances in ensemble learning have promoted various kinds of deep ensemble learning approaches, including using conventional ensemble methods \cite{ref17,ref23} and time-based ensemble methods \cite{ref42}. Among these ensemble learning methods, the decomposition-based ensemble is a less explored category that transforms time-series into simple components for modelling \cite{ref31}. The components extracted from decomposition-based methods may reduce the complexity of the data, and thus yield a more robust solution. However, there is a lack of comparative studies on whether this approach can better benefit deep learning models than non-decomposition-based ensemble methods. Currently, the comparison studies are mostly restricted to certain types of methods \cite{ref31,ref39,ref41}. 

This paper compares several decomposition-based ensemble methods with conventional bagging and time domain multi-resolution ensemble methods \cite{ref42} under the traffic flow forecasting tasks. The main contributions of this paper are:
\begin{enumerate}
\item{An empirical study is conducted to assess the helpfulness of decomposition-based ensemble methods for deep learning models in traffic flow forecasting tasks. Results yield that decomposition-based methods better enhance the performance of deep learning models than baseline methods.}
\item{We explored the effectiveness of optimised aggregation for decomposition methods. The results show that decomposition methods are sensitive to the aggregation methods.}
\item{We investigated the impact of inputs and forecasting horizons for decomposition methods. Results indicate that these methods are sensitive to the input and do not always benefit from extensive data.}
\end{enumerate}
    
\section{Background and related works}
This section briefly reviews the existing decomposition-based ensemble methods and time-domain multi-resolution ensemble approaches.

\subsection{Decomposition-based ensemble methods}
Most research on decomposition-based ensemble methods follows a divide-and-conquer concept, which transforms complex original time-series data into a set of simple components \cite{ref31}. The first category, including Fourier Transform (FT) \cite{ref32} and Wavelet Transform (WT) \cite{ref34}, uses use predefined basis functions. However, predefined basis functions suffer from frequency resolution issues, which means they lack or have a limited ability (due to predefined function) to distinguish between closely spaced frequency components in a signal. To address this gap, the second category avoids predefined functions, extracting signals based on local time scales and, as such, distinguishing closely spaced frequency components into Intrinsic Mode Functions (IMFs). Notable models in this category include Empirical Mode Decomposition (EMD), Ensemble Empirical Mode Decomposition (EEMD), and Complete Ensemble Empirical Mode Decomposition with Adaptive Noise (CEEMDAN) \cite{ref31}. Previous studies \cite{ref36,ref39} show that these methods outperform direct raw data predictions. Thus, we selected EMD, EEMD, and CEEMDAN for this study.

\subsection{Time domain multi-resolution ensemble}
The time interval (i.e., temporal resolution) of recorded time-series data is connected to forecasting performance, making it a beneficial approach for ensemble learning \cite{ref42,ref43}. The challenge of overfitting in deep learning methods due to complex time-series data can be mitigated by considering outcomes of other deep learning models trained on aggregated time-series data that focus on different trends. Previous research suggests that such a multi-resolution ensemble may outperform the single model in terms of performance and robustness \cite{ref42,ref44}.

Frequency resolution refers to the ability to distinguish between closely spaced frequency components in a signal.
\begin{figure}[!ht]
  \centering
  \includegraphics[width=0.9\textwidth]{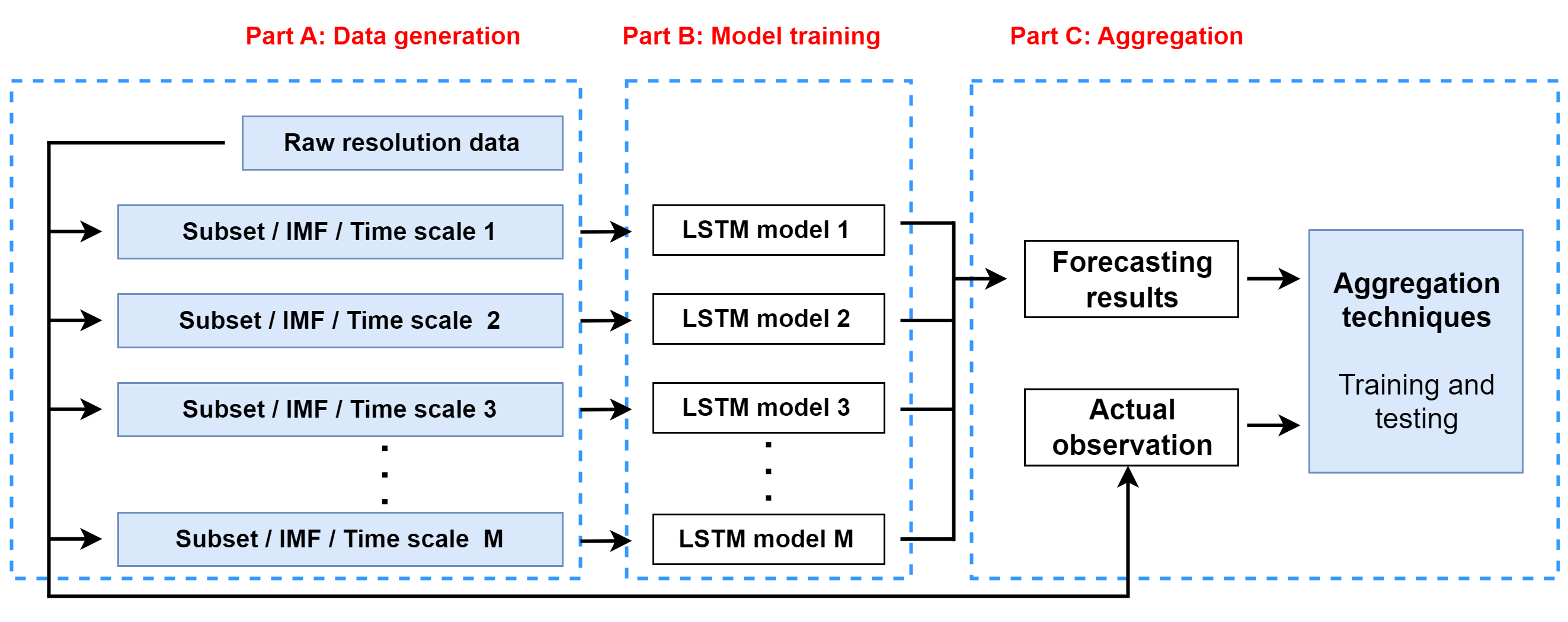}
   \caption{The comparison design with three modules: (a) The data generation module is responsible for extracting the subsets of time patterns from the original time-series. (b) The training module is responsible for training base learners. (c) The aggregating module trains the final learner based on meta-data extracted from base learners.}
   \label{fig:fig1}
\end{figure}

\section{Comparative design}

In this work, we attempt to compare and evaluate the deep ensemble methods on traffic flow forecasting tasks. The proposed comparison design is illustrated in Fig.~\ref{fig:fig1}. 

\subsection{Data generation module}
Firstly, the data generation module extracts the subsets of time patterns from the original time-series for each selected ensemble method. The generated subsets are adjusted according to the specific ensemble learning method. 

\subsubsection{For multi-resolution ensemble and bagging}
The data generation for time domain multi-resolution ensemble is an addition to simple data aggregation, representing views based on aggregation levels \cite{ref46}. Firstly, the original time-series will be sliced into a set of input-output pairs, denoted by set $X = \{ x_{1}, x_{2},..., x_{i} \}$ and set $Y = \{ y_{1}, y_{2},..., y_{i} \}$. Each $x_{i}$ contains $I$ element and each $y_{i}$ is the sum of $T$ elements following $I$. $X^{M} = \{ X^{1}, X^{2},...., X^{m} \}$ is obtained by sum aggregation on $X$ based on each time interval. Assuming the original time-series is recorded in 1 min time interval, input $I$ is 60 minutes, the timer-series input size for $x_{1}^{1}$ in $X^{1}$ and $x_{1}^{10}$ in $X^{10}$ are 60/1 = 60 and 60/10 = 6, respectively. For conventional bagging, the $X^{M}$ and $Y^{M}$ are generated by taking 90\% elements from $X$ and $Y$ with replacement in their original form, and the number of generated series $M$ is equal to 25 as suggested by \cite{ref18}, representing subsets of the original data. 

\subsubsection{For decomposition-based methods}
Since decomposition will alter the feature characteristics of the original time-series data, the input and output are combined into a single sequence $s$ and are sliced after decomposition. The first step is to prepare the raw set $ S = \{ s_{1}, s_{2},..., s_{i} \}$ and each sequence $s_{i}$ contains $I + T$ elements. $ S = \{ s_{1}, s_{2},..., s_{i} \}$ will be decomposed to $M$ sets $\widehat{S^{M}} = \{ \widehat{S^{1}}, \widehat{S^{2}},...., \widehat{S^{m}} \}$ with each $\widehat{S^{m}} = \{ \widehat{s_{1}^{m}}, \widehat{s_{2}^{m}},..., \widehat{s_{i}^{m}} \}$.

After decomposition, we need to extract the input and output from $\widehat{S^{M}}$. Assuming $I$ and $T$ are 60 minutes and 10 minutes, arbitrary $\widehat{s_{i}^{m}}$ in $\widehat{S^{m}}$ can be sliced into $\widehat{x_{i}^{m}} = \{ \widehat{s_{1,i}^{m}},...., \widehat{x_{60,i}^{m}} \}$ and $\widehat{y_{i}^{m}} = \{ \widehat{s_{61,i}^{m}},...., \widehat{s_{70,i}^{m}} \}$ and the task is to forecast $\widehat{y_{i}^{m}}$ using corresponding $\widehat{x_{i}^{m}}$. The final prediction $y_{i}^{'}$ will be aggregated from predicted $\{\widehat{y_{i}^{1'}}, \widehat{y_{i}^{2'}},..., \widehat{y_{i}^{m'}}\}$ and compared with grand truth $y_{i}$. The details of each method are explained below.

\subsubsection{EMD method}
was developed to adaptively decompose complex signals into multiple simpler time-series called Intrinsic Mode Functions (IMFs), providing a better ability to distinguish frequency components compared with the Fourier Transform. We use $s(t)$ for the value in time $t$ in sequence $s$ for simplification. Its process can be described as following steps:

Step (1): local maxima and minima for a sequence $s$ will be determined over time. All local maxima compose an upper envelope $u(t)$ (usually by cubic interpolation), and local minima points will form a lower envelope $l(t)$ in the same way. We calculate the mean value of the upper and lower envelops for each $t$ by:
\begin{equation}
\label{equ_1}
m(t) = (u(t) + l(t))/2.
\end{equation}

Step (2): The $m(t)$ will create a new time-series $m_{envelop}$, and we can get the value of the candidate IMF in each $t$ by
\begin{equation}
\label{equ_2}
IMF_{candidate}(t) = s(t) - m(t)
\end{equation}

Step (3): we need to verify this candidate IMF by checking two conditions: (1) the number of extreme points and the number of zero-crossing points must be equal or differ at most by 1, and (2) the $m_{envelop}$ must be 0 everywhere \cite{ref35}. If these two conditions are not satisfied, steps (1)-(2) must repeatedly apply to this $IMF_{candidate}$ until it meets the criteria. 

When an IMF is found, we will collect it, and the remaining signal becomes:
\begin{equation}
\label{equ_3}
r(t) = s(t) - IMF(t)
\end{equation}

where the sequence $r$ obtained from $r(t)$ will replace the $s$ as input for steps (1)-(3) to extract the next IMF components until the total number of local maxima and minima for residue sequence $r$ can no longer exceed two. The final EMD output is $\{IMF_{1}, IMF_{2},...., IMF_{m}\}$ for given $s$, while the number $m$ of IMFs is not predefined but discovered from the signal. 

\subsubsection{EEMD method}
Some signal components can occasionally appear and disappear within a signal at irregular intervals. The IMF generated by EMD under such a signal can be misleading: (1) several IMFs can be similar, and (2) a single IMF can falsely contain components of different frequencies. Instead of calculating IMF directly, EEMD computes each IMF based on an ensemble of multiple noise-added IMFs. The added white noise provides a uniform reference in the time-frequency space, making the method less sensitive to small variations in the signal that might cause misleading IMFs in standard EMD. Much like the EMD, its process can be described as following steps:

Step (1): Add different Gaussian noise $\omega^{k} (t) , k = 1, 2, . . . ,K $ and corresponding noise coefficient $\epsilon$ to the sequence $s$ to generate the noise-added data $s^{k}(t) = s(t) + \epsilon * \omega^{k} (t), k = 1, 2, . . . ,K$, where $K$ is the number of trials pre-defined. 

Step (2): Each sequence $s^{k}, k = 1, 2, . . . ,K$ is decomposed by EMD to extract all IMFs, i.e., $IMF_{m}^{k}$ refers to $m_{th}$ IMF for $s^{k}$.

Step (3): Get each final IMF for $s$ by ensemble IMFs extracted from $\{ s^{1}, s^{2},.., s^{k} \}$:
\begin{equation}
\label{equ_6}
\overline{IMF_{m}(t)} = \frac {1}{K} \sum_{k=1}^{K} IMF_{m}^{k}(t)
\end{equation}

\subsubsection{CEEMDAN method}
The noise EEMD used improves the IMF quality and hampers the reconstruction ability. Thus, the EEMD does not guarantee an accurate reconstruction of $s$, which can be important. CEEMDAN changes Gaussian noise to adaptive noise, which adds adaptive noise at each decomposition stage and helps to reduce the noise in the final IMFs \cite{ref38}. It ensures a complete decomposition, where the sum of all IMFs and the residual exactly reconstructs the original signal. The quality of IMF is also improved due to an optimal solution for noise addition. Its process can be described as following steps:

Step (1): Add different Gaussian noise $\omega^{k}(t) , k = 1, 2, . . . ,K $ and corresponding noise coefficient $\epsilon_{0}$ to the sequence $s$ to generate the noise-added data $ s^{k}(t) = s(t) + \epsilon_{0} * \omega^{k}(t), k = 1, 2, . . . ,K$, where $K$ is the number of trials pre-defined. 

Step (2): Each $ s^{k}, k = 1, 2, . . . , K$ is decomposed by EMD to extract the first IMF, i.e., $IMF_{1}^{k}$.

Step (3): Calculate the $ \overline{IMF_{1}}$ of $s$ by ensemble the $IMF_{1}^{k}$ using mean values as Eq.~\ref{equ_6}.

Step (4): Calculate the first residual sequence $r_{1}$ like Eq.~\ref{equ_3} by:
\begin{equation}
\label{equ_7}
r_{1}(t) = s(t) - \overline{IMF_{1}(t)} 
\end{equation}

Step (5): we define operator $E(.)$ as an EMD process, and $E_{1}(.)(t)$ represents the value of EMD process's first IMF in time $t$, Then:
\begin{equation}
\label{equ_8}
\overline{IMF_{2}(t)} = \frac {1}{K} \sum_{k=1}^{K} E_{1}(r_{1} + \epsilon_{1} * E_{1}(\omega^{k}))(t)
\end{equation}

Step (6): we further define $E_{m}(.)$, representing EMD processs's $m_{th}$ IMF, then we can acquire all IMFs by:
\begin{equation}
\label{equ_9}
\overline{IMF_{m+1}(t)} = \frac {1}{K} \sum_{k=1}^{K} E_{1}(r_{m} + \epsilon_{m} * E_{m}(\omega^{k}))(t)
\end{equation}

The stop condition is much like the EMD: until the total number of local maxima and minima for residue sequence $r$ can no longer exceed two. Although $\epsilon_{m}$ allows optimisation at each stage of the decomposition, some studies suggest the constant value is also practical \cite{ref38}. 

\subsection{Modelling module}
Long short-term memory (LSTM) is a recurrent neural network type that stores information across the time stamp to model the long-term dependencies \cite{ref47} effectively. We apply LSTM as the base learner in the modelling module for its good time-series learning capacity. Each base learner will be trained in sequential order under the corresponding input-output pair until the maximum training iteration is reached or the early stopping rule on the validation set is violated. A hyperparameter tuning process is responsible for optimising the base learner. To maintain a fair comparison, the input and forecasting horizons are the same for all ensemble learning methods regardless of the data generation process. 

\subsection{Aggregation module}
In the aggregation module, the outputs of base learners will be extracted and utilised using an optimised final learner or baseline aggregation strategy. Mean aggregation is the baseline strategy for time domain multi-resolution ensemble and conventional bagging. Due to the nature of decomposition, sum aggregation is the baseline strategy for decomposition-based methods. We further apply linear regression and MLP as the optimised final learners to utilise the predictions generated from base learners. 

\section{Experimental study}
To evaluate the performance of decomposition-based deep ensemble learning methods, we conducted a series of experiments on Melbourne traffic datasets and PEMS datasets to investigate:
\begin{enumerate}
\item{RQ1: Whether decomposition-based methods better benefit the deep learning models than other ensemble methods on forecasting tasks?}
\item{RQ2: Is the performance of decomposition-based ensemble learning sensitive to aggregation methods?}
\item{RQ3: Whether the answer of RQ1 and RQ2 depends on input and forecasting horizon?}
\end{enumerate}

\begin{figure}[!ht]
  \centering
   \includegraphics[width=0.6\linewidth]{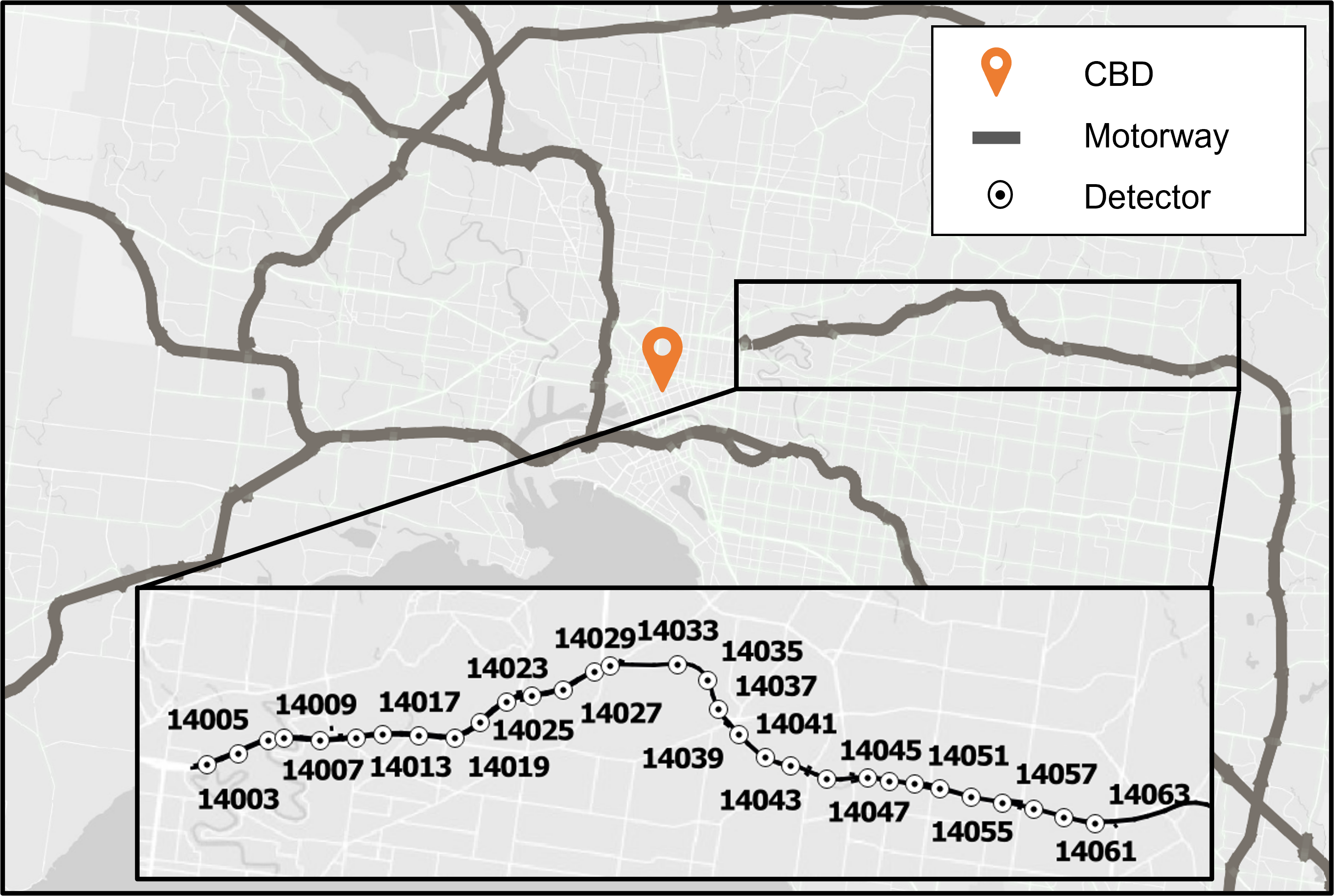}
   \caption{The location of sensors in eastern residential suburbs, Melbourne CBD.}
   \label{fig:fig3}
\end{figure}


\subsection{Data}
In this study, we conducted the experiments on 3 datasets: the Melbourne data set, the PEMSD4 data set and the Portland data set. The Melbourne data set are corridors stretching from Melbourne CBD to the eastern residential suburbs; see Fig.~\ref{fig:fig3}. The traffic flow, speed and occupancy are recorded in 1-minute intervals from 1st Jul 2016 to 30th Sep 2016. In this study, 5 sites that do not have large maintenance (e.g., road construction and fixing) are selected 
regardless of the weekday and weekend.

The collected PEMS data set records the highway traffic flow of the San Francisco Bay Area in California, USA \cite{ref49} from 1st Jan 2018 to 28th Feb 2018 in 30-second intervals. The collected Portland data set comes from the FHWA research project \cite{Portland} in Portland I-205N from September 15, 2011 to November 15, 2011. In each dataset, 5 sites are selected using the same criteria as the Melbourne dataset. For all datasets, the train-val-test set is split on 70\%/10\%/20\%. The validation set is only used for hyperparameter tuning. 

\subsection{The hyper-parameter setting}
To generate IMFs for decomposition-based methods, we use the implementation from the pyEMD package. The Spline method is set to Akima for complex signal processing for all decomposition-based methods. EMD adaptively decompose the signal without preset parameters. For EEMD and CEEMDAN, all noise coefficients $\epsilon$ are set to 0.2, and the number of trials is set to 25. We use the top 5 IMF components in modelling.

The proposed base learners’ model structure is shown in Table~\ref{tab:table1}. LSTM first completes the time-series feature extraction, and the returned sequence is then passed to fully connected layers. The proposed model is implemented using tensorflow-gpu (version 2.6.0), and the hyper-parameter tuning is conducted by keras-tunner (version 1.1.3).

\begin{table}[!ht]
\caption{Hyper-parameter tuning of LSTM model (deep base learner)}
\centering
\label{tab:table1}
\begin{tabular}{|l|l|}
\hline
Model layers & Hidden units                     \\ \hline
LSTM         & {[}8,16,32,64,128{]}             \\ \hline
Dense        & {[}8,16,32,64,128{]}                 \\ \hline
Dropout      & {[}0, 0.1, 0.2, 0.3, 0.4, 0.5{]} \\ \hline
Dense        & {[}8,16,32,64,128{]}                 \\ \hline
Dropout      & {[}0, 0.1, 0.2, 0.3, 0.4, 0.5{]} \\ \hline
Dense        & 10                               \\ \hline
Dense        & Output                               \\ \hline
\end{tabular}
\end{table}

The proposed MLP final learners’ model structure has three dense layers and two dropouts in between. The search space for the dense layer's hidden unit is [8,16,32,64], and the search space for the dropout rate is [0, 0.1, 0.2, 0.3, 0.4, 0.5]. The input and output of the final learner are both 2-dimensional arrays.


\subsection{Evaluation matrix}
This research uses root mean squared error (RMSE) for performance evaluation. RMSE is used to measure the fitness between the predictions and the real observation of traffic flows:
\begin{align*}
RMSE &= \sqrt{\frac {1}{n} * \sum_{i=1}^{n} (tfp_{i} - y_{i})^{2}}  \\
\end{align*}
Where $y_{i}$ is the actual value of traffic flows, $tfp_{i}$ represents the forecasting value corresponding to each $y_{i}$, and $n$ denotes the number of observed values.

\begin{table*}[!t]
\caption{Results of the ensemble learning methods using baseline aggregation against the baseline methods in Melbourne data. The method(s) highlighted in bold if its performance is statistically the best one(s).}
\centering
\label{tab:table3}
\resizebox{\columnwidth}{!}{%
\begin{tabular}{llllll}
\hline
\textbf{}      & \textbf{site14005} & \textbf{site14049} & \textbf{site14059} & \textbf{site14061} & \textbf{site14063} \\ \hline
\textbf{Model} & { \textbf{RMSE}} & { \textbf{RMSE}} & { \textbf{RMSE}} & { \textbf{RMSE}} & { \textbf{RMSE}} \\ \hline
\multicolumn{6}{l}{I   = 120 minutes}                                                                         \\
LSTM                & 59.571          & 90.093          & 57.492          & 56.638          & 47.474          \\
Bagging +   mean    & 48.641          & 61.146          & 45.595          & 47.933          & 36.349          \\
Multi-reso   + mean & 49.031          & 65.619          & 51.342          & 53.779          & 39.157          \\
EMD + sum           & \textbf{25.495} & 41.054          & 36.986          & 35.786          & 33.445          \\
EEMD + sum          & 34.242          & \textbf{30.374} & \textbf{33.360} & 46.221          & \textbf{27.455} \\
CEEMDAN +   sum     & 28.040          & 37.329          & 44.193          & \textbf{35.428} & 29.592          \\ \hline
\multicolumn{6}{l}{I   = 240 minutes}                                                                         \\
LSTM                & 54.201          & 72.146          & 55.417          & 55.770          & 39.333          \\
Bagging +   mean    & 49.042          & 55.783          & 47.451          & 49.639          & 36.745          \\
Multi-reso   + mean & 47.799          & 60.911          & 49.050          & 49.571          & 37.377          \\
EMD + sum           & 32.267          & 46.931          & 36.186          & 33.227          & 31.250          \\
EEMD + sum     & \textbf{31.110}     & \textbf{29.354}     & \textbf{22.904}     & \textbf{30.757}     & \textbf{15.993}     \\
CEEMDAN +   sum     & 31.853          & 40.098          & 38.695          & 30.779          & 27.626          \\ \hline
\multicolumn{6}{l}{I   = 360 minutes}                                                                         \\
LSTM                & 58.276          & 76.819          & 54.426          & 54.874          & 44.023          \\
Bagging +   mean    & 49.079          & 56.300          & 46.259          & 48.500          & 37.524          \\
Multi-reso   + mean & 49.396          & 60.222          & 46.547          & 50.556          & 38.976          \\
EMD + sum           & \textbf{23.204} & 48.403          & 35.376          & 31.975          & \textbf{24.332} \\
EEMD + sum          & 27.552          & \textbf{30.033} & \textbf{31.024} & \textbf{27.598} & 25.395          \\
CEEMDAN +   sum     & 46.705          & 48.293          & 45.486          & 47.839          & 26.736          \\ \hline
\end{tabular}%
}
\end{table*}

\begin{table}[!t]
\caption{Results of the decomposition-based ensemble learning methods using optimised strategy against the baselines aggregation in Melbourne data. The method(s) highlighted in bold if its performance is statistically the best one(s).}
\centering
\label{tab:table4}
\begin{tabular}{llllll}
\hline
\textbf{}      & \textbf{site14005} & \textbf{site14049} & \textbf{site14059} & \textbf{site14061} & \textbf{site14063} \\ \hline
\textbf{Model} & { \textbf{RMSE}} & { \textbf{RMSE}} & { \textbf{RMSE}} & { \textbf{RMSE}} & { \textbf{RMSE}} \\ \hline
\multicolumn{6}{l}{I   = 120 minutes}                                                                        \\
EMD + sum          & 25.495          & 41.054          & 36.986          & 35.786          & 33.445          \\
EMD +   linear     & 24.207          & 33.555          & 31.979          & 31.470          & 28.858          \\
EMD + MLP          & 27.685          & 38.204          & 34.679          & 36.811          & 35.994          \\
EEMD + sum         & 34.242          & 30.374          & 33.360          & 46.221          & 27.455          \\
EEMD +   linear    & \textbf{22.142} & \textbf{26.747} & \textbf{22.730} & \textbf{21.193} & \textbf{17.567} \\
EEMD + MLP         & 23.975          & 27.544          & 23.122          & 24.551          & 22.321          \\
CEEMDAN +   sum    & 28.040          & 37.329          & 44.193          & 35.428          & 29.592          \\
CEEMDAN +   linear & 26.990          & 31.893          & 35.267          & 33.211          & 27.943          \\
CEEMDAN +   MLP    & 27.632          & 29.299          & 40.161          & 32.815          & 28.029          \\ \hline
\multicolumn{6}{l}{I   = 240 minutes}                                                                        \\
EMD + sum          & 32.267          & 46.931          & 36.186          & 33.227          & 31.250          \\
EMD +   linear     & 26.184          & 31.717          & 31.878          & 29.021          & 30.491          \\
EMD + MLP          & 37.696          & 33.864          & 35.806          & 33.640          & 31.292          \\
EEMD + sum         & 31.110          & 29.354          & 22.904          & 30.757          & 15.993          \\
EEMD +   linear    & \textbf{20.579} & \textbf{25.382} & \textbf{21.272} & \textbf{22.127} & \textbf{15.363} \\
EEMD + MLP         & 23.927          & 25.726          & 21.505          & 23.093          & 25.374          \\
CEEMDAN +   sum    & 31.853          & 40.098          & 38.695          & 30.779          & 27.626          \\
CEEMDAN +   linear & 28.842          & 31.542          & 30.484          & 29.071          & 23.270          \\
CEEMDAN +   MLP    & 30.957          & 35.395          & 29.155          & 37.400          & 30.473          \\ \hline
\multicolumn{6}{l}{I   = 360 minutes}                                                                        \\
EMD + sum          & 23.204          & 48.403          & 35.376          & 31.975          & 24.332          \\
EMD +   linear     & 19.965          & 33.480          & 28.697          & 30.292          & 22.381          \\
EMD + MLP          & 24.914          & 43.897          & 35.089          & 33.223          & 22.920          \\
EEMD + sum         & 27.552          & 30.033          & 31.024          & 27.598          & 25.395          \\
EEMD +   linear    & \textbf{18.072} & \textbf{26.038} & \textbf{21.379} & \textbf{22.792} & \textbf{18.165} \\
EEMD + MLP         & 23.655          & 28.645          & 22.166          & 25.464          & 18.232          \\
CEEMDAN +   sum    & 46.705          & 48.293          & 45.486          & 47.839          & 26.736          \\
CEEMDAN +   linear & 28.026          & 38.748          & 40.342          & 37.731          & 20.284          \\
CEEMDAN +   MLP    & 43.151          & 39.521          & 45.883          & 42.264          & 29.699          \\ \hline
\end{tabular}

\end{table}

\subsection{Overall comparison}
To assess the effectiveness of decomposition-based ensemble methods, we compared EMD, EEMD, and CEEMAND against conventional bagging and multi-resolution ensemble, all trained on time-series recorded in 1-minute intervals over 10 repeated runs. A standalone LSTM model, trained on the aggregated time-series data with an equal time interval and forecasting horizon, was also compared with the aforementioned methods as an additional baseline, as suggested by \cite{ref46}. The input horizon consists of 120, 240, and 360 minutes, with a forecasting horizon of 10 minutes for this test. The results for the Melbourne dataset are reported in Table~\ref{tab:table3}, while the other results are reported in the supplementary material \footnote[1]{\href{https://alexkaiqin.org/doc/AJCAI2024DDELSupp.pdf}{\label{note1} https://alexkaiqin.org/doc/AJCAI2024DDELSupp.pdf}}. Based on RMSE measurements, EEMD is the best ensemble in the Melbourne dataset, outperforming other methods in 10 out of 15 test scenarios. In the case of the PEMS and Portland datasets, EEMD outperformed other methods in 13 and 14 test scenarios, respectively. Overall, EEMD methods outperform the conventional bagging and multi-resolution ensemble methods. In all cases, decomposition-based methods outperformed non-decomposition-based methods. 
\subsection{The ablation study of aggregation strategy}
We compare linear regression and MLP methods against baseline aggregation across the ensemble methods to assess the sensitivity of decomposition-based methods to aggregation strategy. The test setup is the same as the overall comparison. The results for the Melbourne dataset are reported in Table~\ref{tab:table4}, and the other results are reported in the supplementary material. Based on RMSE measurement, linear is the best ensemble in the Melbourne dataset. In the case of the PEMS and Portland datasets, the best strategy is also linear. 

Furthermore, linear regression and MLP outperform the baseline aggregation strategy in 127 and 94 cases, respectively. This might be because our progress didn't add residue to modelling, and some uncovered IMFs remain out of the top 5 IMFs, which are further utilised by linear regression and MLP methods.

\begin{figure*}
  \begin{subfigure}[t]{.45\textwidth}
    \centering
    \includegraphics[width=\linewidth]{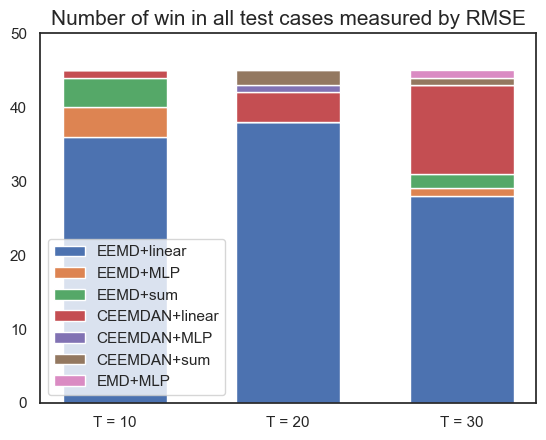}
    \subcaption{Number of wins for all methods in all test cases measured by RMSE.}\label{fig:fig6}
  \end{subfigure}
  \hfill
  \begin{subfigure}[t]{.48\textwidth}
    \centering
    \includegraphics[width=\linewidth]{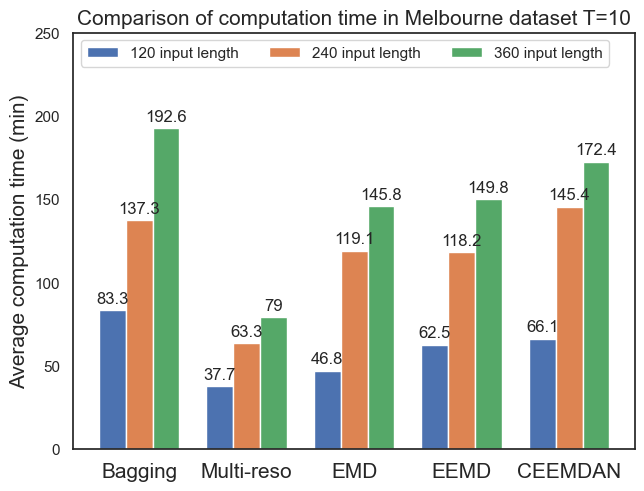}
    \subcaption{Comparison of computation time (in minutes) of conventional bagging, multi-resolution ensemble, EMD, EEMD, and CEEMDAN methods in Melbourne dataset, target horizon 10.}\label{fig:fig5}
  \end{subfigure}

\caption{The computation time analysis and sensitivity analysis.}
\end{figure*}

\subsection{Sensitivity of 20-min and 30-min forecasting horizons}
We further evaluate the performance of all combinations of ensemble and aggregation methods for the forecasting tasks with 20-min and 30-min forecasting horizons. The rest of the test setup is the same as the overall comparison. The results are summarised in Fig~\ref{fig:fig6}. We evaluated the results and noticed that 360-min input would improve the general performance of non-decomposition-based methods. This is consistent with the previous study \cite{input_effect} that longer inputs generally improve forecasting accuracy. However, decomposition-based are not sensitive to specific setups, and the CEEMDAN method tends to perform better when the output size is large. Fig.~\ref{fig:fig5}. compares each ensemble's computation time (in minutes) paired with linear regression for different inputs. It can be observed that decomposition-based methods have higher computation time due to the IMF construction progress, whereas CEEMDAN is more computationally expensive because of the adaptive noise strategy.

\section{Conclusions and Future Work}
In this paper, we compare three decomposition-based deep ensemble learning methods with two common non-decomposition-based ones, including bagging and multi-resolution ensemble, for their performance in solving traffic flow forecasting tasks. Experimental results on several traffic flow datasets demonstrate the superiority of decomposition-based methods, with the EEMD-based method outperforming others in most test scenarios. Future work includes the exploration of advanced ensemble learning strategies based on multi-task optimisation \cite{xu2022evolutionary} and the incorporation of other types of base learners \cite{qin2005initialization}. We also plan to evaluate the decomposition-based methods in other application scenarios involving time-series forecasting missions \cite{mistry2018meta}.

\begin{credits}


\subsubsection{\ackname} This work was supported by the Australian Research Council (ARC) under Grant No. LP180100114. During the preparation of this work the authors used ChatGPT in order to improve language. After using this tool, the authors reviewed and edited the content as needed and take full responsibility for the content of the publication.

\end{credits}
%
%
%

\bibliographystyle{splncs04}
\bibliography{main}

\begin{thebibliography}{10}
\providecommand{\url}[1]{\texttt{#1}}
\providecommand{\urlprefix}{URL }
\providecommand{\doi}[1]{https://doi.org/#1}

\bibitem{ref46}
Abdulhai, B., Porwal, H., Recker, W.: Short-term traffic flow prediction using neuro-genetic algorithms. Intelligent Transportation Systems Journal  \textbf{7}(1),  3–41 (Jan 2002). \doi{10.1080/10248070212011}, \url{http://dx.doi.org/10.1080/10248070212011}

\bibitem{ref49}
Bai, L., Yao, L., Li, C., Wang, X., Wang, C.: Adaptive graph convolutional recurrent network for traffic forecasting. ArXiv  \textbf{abs/2007.02842} (2020), \url{https://api.semanticscholar.org/CorpusID:220363737}

\bibitem{ref42}
Chen, C., Liu, H.: Medium-term wind power forecasting based on multi-resolution multi-learner ensemble and adaptive model selection. Energy conversion and management  \textbf{206},  112492 (2020)

\bibitem{ref32}
Chen, L., Zheng, L., Yang, J., Xia, D., Liu, W.: Short-term traffic flow prediction: From the perspective of traffic flow decomposition. Neurocomputing  \textbf{413},  444--456 (2020)

\bibitem{ref18}
Chen, L., Chen, C.P.: Ensemble learning approach for freeway short-term traffic flow prediction. In: 2007 IEEE International Conference on System of Systems Engineering. pp.~1--6. IEEE (2007)

\bibitem{ref34}
Chen, S., Wang, W.: Traffic volume forecasting based on wavelet transform and neural networks. In: International Symposium on Neural Networks. pp.~1--7. Springer (2006)

\bibitem{ref6}
Hansen, L.K., Salamon, P.: Neural network ensembles. IEEE transactions on pattern analysis and machine intelligence  \textbf{12}(10),  993--1001 (1990)

\bibitem{ref47}
Hochreiter, S., Schmidhuber, J.: Long short-term memory. Neural Computation  \textbf{9}(8),  1735–1780 (Nov 1997). \doi{10.1162/neco.1997.9.8.1735}, \url{http://dx.doi.org/10.1162/neco.1997.9.8.1735}

\bibitem{ref41}
Huang, H., Chen, J., Huo, X., Qiao, Y., Ma, L.: Effect of multi-scale decomposition on performance of neural networks in short-term traffic flow prediction. IEEE access  \textbf{9},  50994--51004 (2021)

\bibitem{ref35}
Huang, N.E., Shen, Z., Long, S.R., Wu, M.C., Shih, H.H., Zheng, Q., Yen, N.C., Tung, C.C., Liu, H.H.: The empirical mode decomposition and the hilbert spectrum for nonlinear and non-stationary time series analysis. Proceedings of the Royal Society of London. Series A: mathematical, physical and engineering sciences  \textbf{454}(1971),  903--995 (1998)

\bibitem{ref44}
Liu, H., Duan, Z., Chen, C.: Wind speed big data forecasting using time-variant multi-resolution ensemble model with clustering auto-encoder. Applied Energy  \textbf{280},  115975 (2020). \doi{https://doi.org/10.1016/j.apenergy.2020.115975}, \url{https://www.sciencedirect.com/science/article/pii/S0306261920314252}

\bibitem{ref39}
Lu, W., Rui, Y., Yi, Z., Ran, B., Gu, Y.: A hybrid model for lane-level traffic flow forecasting based on complete ensemble empirical mode decomposition and extreme gradient boosting. IEEE Access  \textbf{8},  42042--42054 (2020)

\bibitem{mistry2018meta}
Mistry, S., Bouguettaya, A., Dong, H., Qin, A.K.: Metaheuristic optimization for long-term iaas service composition. IEEE Transactions on Services Computing  \textbf{11}(1),  131--143 (2018)

\bibitem{ref17}
Moretti, F., Pizzuti, S., Panzieri, S., Annunziato, M.: Urban traffic flow forecasting through statistical and neural network bagging ensemble hybrid modeling. Neurocomputing  \textbf{167}, ~3--7 (2015)

\bibitem{ref3}
Okutani, I., Stephanedes, Y.J.: Dynamic prediction of traffic volume through kalman filtering theory. Transportation Research Part B: Methodological  \textbf{18}(1),  1--11 (1984)

\bibitem{input_effect}
Petelin, G., Hribar, R., Papa, G.: Models for forecasting the traffic flow within the city of ljubljana. European Transport Research Review  \textbf{15}(1), ~30 (2023)

\bibitem{ref2}
Qiao, F., Yang, H., Lam, W.H.: Intelligent simulation and prediction of traffic flow dispersion. Transportation Research Part B: Methodological  \textbf{35}(9),  843--863 (2001)

\bibitem{qin2005initialization}
Qin, A.K., Suganthan, P.N.: Initialization insensitive {LVQ} algorithm based on cost-function adaptation. Pattern Recognition  \textbf{38}(5),  773--776 (2005)

\bibitem{Portland}
RESEARCH, T., CENTER, E.: Multimodal transportation data research, \url{https://trec.pdx.edu/transportation-data-research}, accessed: 26 April 2024

\bibitem{ref31}
Shi, J., Leau, Y.B., Li, K., Park, Y.J., Yan, Z.: Optimization and decomposition methods in network traffic prediction model: A review and discussion. IEEE Access  \textbf{8},  202858--202871 (2020)

\bibitem{ref4}
Tedjopurnomo, D.A., Bao, Z., Zheng, B., Choudhury, F.M., Qin, A.K.: A survey on modern deep neural network for traffic prediction: Trends, methods and challenges. IEEE Transactions on Knowledge and Data Engineering  \textbf{34}(4),  1544--1561 (2020)

\bibitem{ref38}
Torres, M.E., Colominas, M.A., Schlotthauer, G., Flandrin, P.: A complete ensemble empirical mode decomposition with adaptive noise. In: 2011 IEEE international conference on acoustics, speech and signal processing (ICASSP). pp. 4144--4147. IEEE (2011)

\bibitem{ref1}
Vlahogianni, E.I., Karlaftis, M.G., Golias, J.C.: Short-term traffic forecasting: Where we are and where we’re going. Transportation Research Part C: Emerging Technologies  \textbf{43},  3--19 (2014)

\bibitem{ref36}
Wang, H., Liu, L., Dong, S., Qian, Z., Wei, H.: A novel work zone short-term vehicle-type specific traffic speed prediction model through the hybrid emd--arima framework. Transportmetrica B: Transport Dynamics  \textbf{4}(3),  159--186 (2016)

\bibitem{ref9}
Wang, Z.j., Liu, H.x., Qiu, S., Fang, J.p., Wang, T.: The predictability of short-term urban rail demand: choice of time resolution and methodology. Sustainability  \textbf{11}(21), ~6173 (2019)

\bibitem{xu2022evolutionary}
Xu, H., Qin, A.K., Xia, S.: Evolutionary multitask optimization with adaptive knowledge transfer. IEEE Transactions on Evolutionary Computation  \textbf{26}(2),  290--303 (2022)

\bibitem{ref43}
Zhong, C., Batty, M., Manley, E., Wang, J., Wang, Z., Chen, F., Schmitt, G.: Variability in regularity: Mining temporal mobility patterns in london, singapore and beijing using smart-card data. PloS one  \textbf{11}(2),  e0149222 (2016)

\bibitem{ref23}
Zhou, T., Han, G., Xu, X., Lin, Z., Han, C., Huang, Y., Qin, J.: $\delta$-agree adaboost stacked autoencoder for short-term traffic flow forecasting. Neurocomputing  \textbf{247},  31--38 (2017)

\end{thebibliography}

\end{document}